\documentclass{article}
\usepackage{spconf,amsmath,graphicx}

\usepackage{times}
\usepackage{epsfig}
\usepackage{graphicx}
\usepackage{amsmath}
\usepackage{amssymb}
\usepackage{booktabs}
\usepackage{enumitem}
\usepackage{soul}
\usepackage{color}
\usepackage[para]{footmisc}

\newcommand{\keypoint}[1]{\vspace{0.1cm}\noindent\textbf{#1}\quad}
\newcommand{\cut}[1]{}
\newcommand{\etal}{\textit{et. al. }}
\usepackage{algorithm}
\usepackage{algpseudocode}

\usepackage[nopar]{lipsum}
\usepackage{arydshln}
\usepackage{color, colortbl}
\definecolor{LightCyan}{rgb}{0.95,0.95,0.95}
\newcolumntype{a}{>{\columncolor{LightCyan}}c}
\usepackage[breaklinks,colorlinks]{hyperref}

\title{IDEAL: Improved DEnse locAL Contrastive Learning for Semi-Supervised Medical Image Segmentation}
\name{Hritam Basak$^{1}$, Soumitri Chattopadhyay$^{*2}$, Rohit Kundu$^{*2}$, Sayan Nag$^{*3}$, Rammohan Mallipeddi $^4$}
\address{$^1$Stony Brook University, $^2$Jadavpur University, $^3$University of Toronto, $^4$Kyungpook National University}
\begin{document}
%
\maketitle
\begin{abstract}
Due to the scarcity of labeled data, Contrastive Self-Supervised Learning (SSL) frameworks have lately shown great potential in several medical image analysis tasks. However, the existing contrastive mechanisms are sub-optimal for dense pixel-level segmentation tasks due to their inability to mine local features. To this end, we extend the concept of metric learning to the segmentation task, using a dense (dis)similarity learning for pre-training a deep encoder network, and employing a semi-supervised paradigm to fine-tune for the downstream task. Specifically, we propose a simple convolutional projection head for obtaining dense pixel-level features, and a new contrastive loss to utilize these dense projections thereby improving the local representations. A bidirectional consistency regularization mechanism involving two-stream model training is devised for the downstream task. Upon comparison, our IDEAL method outperforms the SoTA methods by fair margins on cardiac MRI segmentation. Our source codes are publicly accessible at: \href{https://github.com/Rohit-Kundu/IDEAL-ICASSP23}{https://github.com/Rohit-Kundu/IDEAL-ICASSP23}.

\end{abstract}

\def\thefootnote{*}\footnotetext{\hspace{-1.5mm}Equal contribution.}\def\thefootnote{\arabic{footnote}}
\section{Introduction}
\label{sec:intro}

The success of supervised deep learning approaches can be attributed to the availability of large quantities of labeled data that is essential for network training \cite{basak2022mfsnet}. However, in the biomedical domain \cite{shurrab2022self, ronneberger2015unet}, it is difficult to acquire such large quantities of annotated data as the annotations are to be done by trained medical professionals. Although supervised learning has been used extensively in the past decade in biomedical imaging \cite{basak2022mfsnet}, Self-supervised Learning (SSL) \cite{doersch2015unsupervised, pathak2016context, chen2020simple, he2020momentum, manna2022swis} provides more traction for sustaining deep learning methods in the medical vision domain \cite{basak2022embarrassingly, chen2019self, taleb20203d}. SSL-based pre-training alleviates the data annotation problem by utilizing only unstructured data to learn distinctive information which can further be utilized in downstream applications, typically in a semi-supervised fashion \cite{basak2022addressing, chaitanya2020contrastive, basak2022embarrassingly}. Typically, SSL have shown to be exceedingly promising in domains with enormous amounts of data such as natural images \cite{chen2020simple, he2020momentum} with a focus on both contrastive \cite{oord2018representation, he2020momentum, chen2020simple, wang2021dense} and non-contrastive variants \cite{zbontar2021barlow, bardes2021vicreg}. Recently, SSL has begun to be employed for medical imaging as well, evident from the survey \cite{shurrab2022self}.

\begin{figure*}[!t]
\begin{center}
  \includegraphics[width=\linewidth]{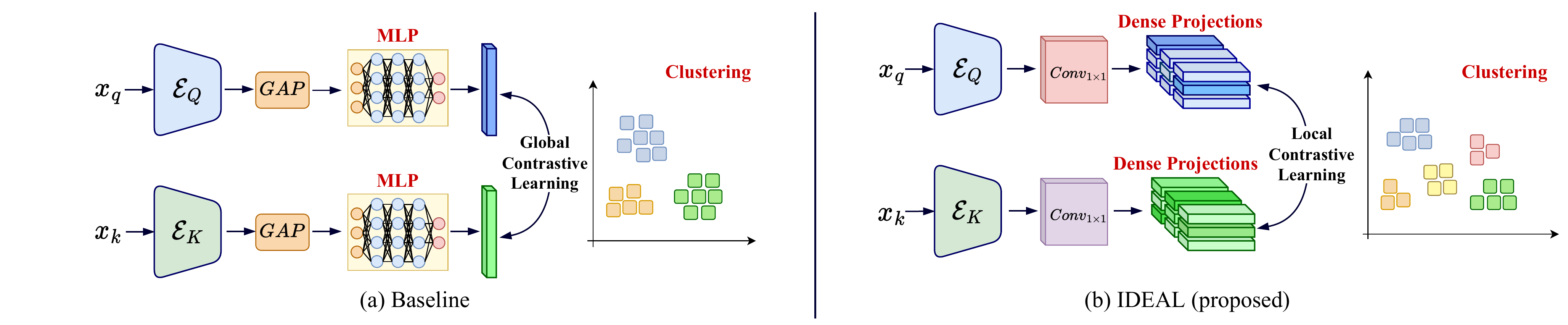}
\end{center}
\vspace{-0.4cm}
 \caption{Schematic representation of contrastive learning-based pre-training of (a) baseline model \cite{chaitanya2020contrastive} (b) our IDEAL model. $\mathcal{E}$ represents the encoder backbone; $GAP =$ Global Average Pooling; $x_q$ and $x_k$ are the query and key images respectively. The convolution projection head in the IDEAL model generates dense projections that give better local clusters than using a traditional global pooling followed by a Multi-Layer Perceptron (MLP).}
\vspace{-0.4cm}
\label{fig:teaser}
\end{figure*}

One of the most popular ways of employing SSL is contrastive learning \cite{oord2018representation, he2020momentum, chen2020simple}, which owing to its prolific performance across various vision tasks has almost become a de facto standard for self-supervision. The intuition of contrastive learning is to \textit{pull} embeddings of semantically similar objects closer and simultaneously \textit{push} away the representations of dissimilar objects. In medical image segmentation, contrastive learning has been leveraged in a few prior works \cite{chaitanya2020contrastive, zeng2021positional}.  However, naive contrastive learning frameworks such as SimCLR \cite{chen2020simple} and MoCo \cite{he2020momentum} learn a global representation and thus, are not suitable to be applied directly to segmentation, a pixel-level task. The seminal work by Chaitanya \etal \cite{chaitanya2020contrastive} utilized a contrastive pre-training strategy to learn useful representations for a segmentation task in a low-data regime. Despite having success, their method has a few limitations, which we attempt to address and improve upon in this paper. In particular, we take a cue from the aforementioned work \cite{chaitanya2020contrastive} to propose a novel contrastive learning-based medical image segmentation framework.

Although our work is built upon \cite{chaitanya2020contrastive}, there are several salient points of difference between the two models that render our contributions non-trivial and unique. First of all, our proposed method preserves spatial information and constructs a dense output projection that preserves locality. This is in contrast to the projection used in \cite{chaitanya2020contrastive} where a global pooling is applied to the backbone encoder, thereby obtaining a single global feature representation vector for every input image. In other words, \cite{chaitanya2020contrastive}  employed global contrastive learning for encoder pre-training and the local contrastive learning is only restricted to the decoder fine-tuning. In contrast, our method involves pre-training of the encoder on dense local features. The intuitive difference between our proposed framework and that of \cite{chaitanya2020contrastive} has been depicted in Fig. \ref{fig:teaser}. We argue that this will benefit pixel-level informed downstream tasks such as segmentation (supported by our findings in Section \ref{sec:expt}). 

Secondly, our definition of finding positive and negative pairs for contrastive learning is different from that of traditional contrastive \cite{chen2020simple, he2020momentum} methods. This is because, in our case, we find dense correspondences across views and define positives and negatives accordingly, i.e., we perform feature-level Contrastive Learning (refer to Section \ref{sec:method}). Further, we adopt a novel bidirectional consistency regularization mechanism that perturbs the network instead of image perturbation used by most frameworks \cite{he2020momentum, zou2020pseudoseg, basak2022embarrassingly}. By this approach, we enforce both streams of data flow to learn better features competitively, unlike the traditional image augmentation-based consistency regularization approaches where only one model stream learns through a single backpropagation. The empirical evaluation suggests that our proposed consistency regularization yields a superior segmentation performance, outperforming existing state-of-the-art semi-supervised approaches in literature \cite{bai2017semi, chaitanya2019semi}. Our contributions are as follows:
\begin{enumerate}
    \item We propose an SSL strategy that leverages dense projection head representations for contrastive learning for learning robust local features. 
    
    \item We redefine 'positive' and 'negative' samples in contrastive learning and extend the InfoNCE loss to adapt to dense representations during the pre-training phase.
    
    \item Our work also employs a uniquely devised cross-consistency regularization for fine-tuning the network to the downstream segmentation task.
\end{enumerate}

\begin{figure*}[tbp]
    \centering
    \includegraphics[width=\linewidth]{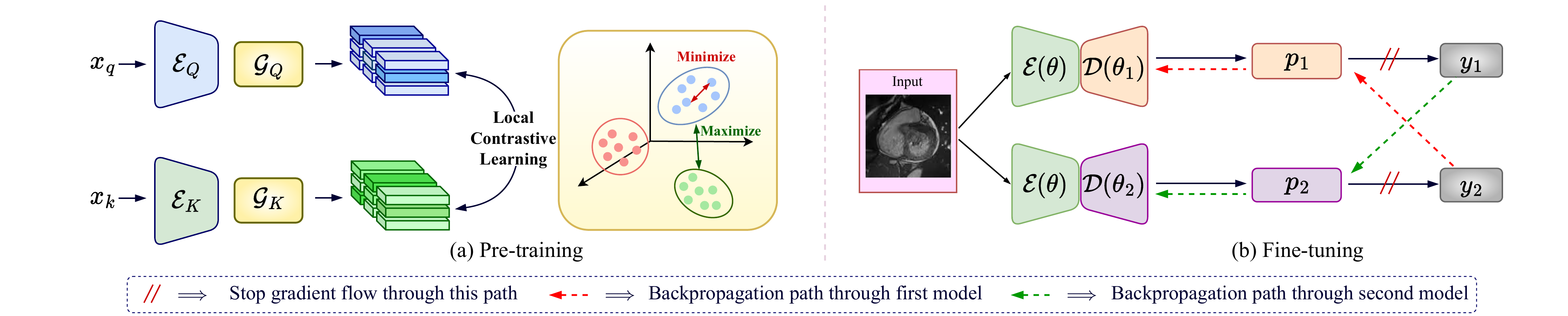}
    \caption{An overview of the proposed IDEAL framework: (a) \textit{Pre-training}- $x_q$ and $x_k$ are the query and key images, $\mathcal{E}$ and $\mathcal{G}$ represent the encoder and projection head, respectively. The projection head employs a 1x1 convolution layer instead of a traditional MLP for dense feature extraction, resulting in better local clustering of features. (b) \textit{Fine-tuning}- Two perturbed branches with the same input are employed. $\mathcal{E}(\theta)$ is the shared encoder initialized similarly for both streams, $\mathcal{D}(\theta_1)$ and $\mathcal{D}(\theta_2)$ represent two different decoder architectures; $p_1$ and $p_2$ are the predicted output segmentation maps which are thresholded to obtain $y_1$ and $y_2$ respectively. $y_1$ backpropagates through the second stream and $y_2$ backpropagates through the first stream enforcing cross-consistency in segmentation.}
    \label{fig:mainfig}
\end{figure*}

\section{Methodology}\label{sec:method}
Our proposed method consists of two parts: (1) Self-Supervised pre-training of the encoder, and (2) Semi-Supervised fine-tuning of the network using consistency regularization. Taking reference from \cite{chaitanya2020contrastive}, we propose a pairwise contrastive (dis)similarity learning, employing global and local feature exploration for pre-training the encoder. The pre-trained encoder is thereafter transferred for the downstream task of cross-supervision-based consistency regularization, which aims to fine-tune the network. 

\subsection{Background: Global Contrastive Learning}\label{sec:GCL}
The widely employed variant of self-supervised representation learning is contrastive learning \cite{he2020momentum, chen2020simple, oord2018representation}, where the images and their perturbations are fed into a cascade of the shared encoder and projection head. The obtained features (global representations) of each of these images in the mini-batches are then utilized for a contrastive loss function, which is designed considering contrastive learning as a dictionary lookup task \cite{wang2021dense}. Out of all the encoded keys in the set $k$ for each encoded query $q$, a single positive key $k^+$ is matched, whilst the remainder of the keys (negative keys) represent different images of the mini-batch. A contrastive loss function is represented as follows:
\begin{equation}
    \mathcal{L}_g = -\log \frac{\exp(q\cdot k^+/\tau)}{\exp(q\cdot k^+/\tau) + \sum\limits_{k^-}\exp(q\cdot k^-/\tau)}
\end{equation}

\subsection{Improved Local Contrastive Learning}\label{sec:LCL}
We employ local contrastive Learning that extends the existing global contrastive framework into a dense paradigm. The key difference between the traditional contrastive framework and the current work lies in the formulation of the loss function and the projection head. The traditional contrastive learning follows the paradigm where features are extracted by a backbone encoder (e.g. ResNet \cite{he2016deep}), followed by a projection head. The projection head typically consists of a global pooling operation and an MLP module comprising a few FC layers with ReLU layers in between them. However, in our case, we remove the entire global pooling and MLP part and replace it with a $1\times1$ convolution operation ($\mathcal{G}$). This projection head generates dense feature representations, with an equal number of parameters as the traditional ones. 

We define a set of keys $\{k_0, k_1, ....\}$ extracted from the encoder for every query $q$, where the encoded keys represent the local part of the image, which is very different from traditional contrastive learning where every key corresponds to a single image view. These keys represent individual features of the $D_h\times D_w$ vectors extracted from the encoder followed by $1\times1$ convolution, where $D_h$ and $D_w$ represent the spatial dimension of the extracted feature maps. The positive keys $k^+$ are pooled from the extracted correspondence across views, i.e. one of the $D_h\times D_w$ features extracted from different views of the same image. The negative key $k^-$ is assigned from features from a view of different images. Having this assignment strategy and taking reference from InfoNCE \cite{oord2018representation}, we define the local contrastive loss as:

\begin{equation}
    \mathcal{L}_{loc} = \frac{-1}{D_h D_w} \sum\limits_{i=1}^{D_h D_w} \log \frac{\exp(q_i\cdot k_i^+/\tau)}{\exp(q_i\cdot k_i^+/\tau) + \sum\limits_{k_i^-}\exp(q_i\cdot k_i^-/\tau)}
\end{equation}

\subsection{Semi-Supervised Fine-tuning}
The SSL pre-training phase is followed by a semi-supervised fine-tuning of the trained encoder network for our downstream segmentation task. Consistency regularization \cite{basak2022embarrassingly, chaitanya2020contrastive}, a very popular semi-supervised approach for segmentation, has been employed here, however, with two very unique and non-trivial customization, which are discussed below. 

In typical consistency regularization schemes, such as in MoCo \cite{he2020momentum}, dampened sharing of weights takes place i.e. one branch is learned directly following the gradient updates whereas the other branch is learned by employing an Exponential Moving Average (EMA) weight transfer strategy. Such a learning paradigm limits the feature learning ability of the second branch. To alleviate this, we consider two decoders of different architecture initialization. Furthermore, the up-scaling convolutions corresponding to each of them are made to work differently, i.e., one of them has ConvTranspose (convolution with trainable kernels) whilst the other one has an Up-sampling layer (simple bilinear interpolation). Outputs of each of these decoders are passed through sigmoid activation functions resulting in $p_i$ which are then threshold-ed to generate one-hot $y_i$ (see Fig. \ref{fig:mainfig}). For labeled samples, we compute a simple cross-entropy loss between target mask $m$ and the predicted outputs $p_i$. Otherwise, a cross-consistency loss is enforced between the output probabilities $p_i$ from the two decoder branches, guided by the one-hot pseudo-label $y_j$ coming from the other branch. In other words, $y_1$ acts as a pseudo-supervisory signal for $p_2$, and $y_2$ for $p_1$. Thus, one branch learns based on the other branch's output, leading to a \emph{competitive learning} between two branches, resulting more confident predictions. The loss function is defined as:

\begin{equation}
\mathcal{L}_{SSF} =
\begin{cases}
CE(p_1, m) + CE(p_2, m),     \text{ for labeled set}\\
CE(p_1, y_2) + CE(p_2, y_1), \text{ for unlabeled set}
\end{cases}
\end{equation}


%

\section{Experiments and Results}\label{sec:expt}
\subsection{Experimental Setup}
\keypoint{Datasets:}\label{sec:data} The proposed method has been evaluated on two publicly available MRI datasets - (1) the ACDC dataset \cite{bernard2018deep}, consisting of 100 cardiac 3D shot-axis MRI volumes with expert manual annotations for three structures, namely, the left and right ventricles and the myocardium; and (2) the MMWHS dataset \cite{zhuang2013challenges, zhuang2016multi}, consisting of 20 cardiac 3D MRI volumes with expert manual annotations for seven structures: the left and right ventricles, the left and right atria, the pulmonary artery, the myocardium, and the ascending aorta. The datasets are split into train, validation, and test sets, where the validation set for each dataset consists of 2 volumes; the test set consists of 20 volumes for the ACDC dataset and 10 volumes for the MMWHS dataset. The rest of the volumes (78 for ACDC and 8 for MMWHS) are used for training. The training and validation set together form the pre-training set, while the testing set was used \emph{solely} for model evaluation.


\keypoint{Implementation Details:}\label{sec:impl} 
IDEAL uses a ResNet-50 \cite{he2016deep} encoder backbone with ADAM optimizer and an initial learning rate of $1e-5$, and a U-Net \cite{ronneberger2015unet} decoder with 4 upscaling layers. The IDEAL model was implemented in PyTorch \cite{paszke2019pytorch} and accelerated by an NVIDIA Tesla K80 GPU.

\keypoint{Evaluation Metrics:}\label{sec:eval} Three different metrics have been used to evaluate the IDEAL model on the segmentation tasks: Dice Similarity Coefficient (DSC), Average Symmetric Distance (ASD), and Hausdorff Distance (HD). The average scores over 5 runs for all the metrics have been reported for all the fine-tuning experiments.

\subsection{Performance Analysis}\label{sec:performance}

\subsubsection{Fine-tuning performances on limited annotations}

\begin{table}[!ht]
    \centering
    \resizebox{\columnwidth}{!}{
    \begin{tabular}{l cccc cccc}

    \toprule
    \textbf{Metric} & \multicolumn{4}{c}{\textbf{ACDC}} & \multicolumn{4}{c}{\textbf{MMWHS}}\\[0.5ex]
    \cmidrule(lr){2-5}
    \cmidrule(lr){6-9}
    & \textbf{L=1.25\%} & \textbf{L=2.5\%} & \textbf{L=10\%} & \textbf{L=100\%} & \textbf{L=10\%} & \textbf{L=20\%} & \textbf{L=40\%} & \textbf{L=100\%} \\[0.5ex]
    \midrule
    
    \textbf{ASD} $\downarrow$  & 0.677	& 0.614	& 0.556	& 0.489	& 2.397	& 1.798	& 1.355	& 1.317 \\[.5ex]
    
    \textbf{HD} $\downarrow$ & 2.409 & 2.209 & 2.094 & 1.999 & 5.001 & 3.499 & 2.221 & 2.183 \\[.5ex]

    \textbf{DSC} $\uparrow$ & 0.738 & 0.846 & 0.879 & 0.882 & 0.626 & 0.791 & 0.815 & 0.826 \\[.5ex]
    
    \bottomrule
    
    \end{tabular} %
    }
    \caption{Results obtained by the IDEAL framework with varying amounts of labeled data on the ACDC and MMWHS datasets. `$L$' represents the amount of labeled data used.}
    \label{tab:metrics}
\end{table}

To study the segmentation performance under limited labels during the fine-tuning phase, we have experimented with 1.25\%, 2.5\%, and 10\% labeled volumes (denoted by `$L$' henceforth) for ACDC and 10\%, 20\%, and 40\% labeled volumes for the MMWHS dataset, following the existing works in literature that have used the same datasets \cite{basak2022embarrassingly, chaitanya2020contrastive}. The amount of labeled data is expressed as a percentage of the training set. The results are tabulated in Table \ref{tab:metrics}. From the results, it can be inferred that on both datasets, the expected trend of improving metrics with an increase in labeled data for the fine-tuning phase. Further, it is promising to observe that even with a fraction of labels available (10\% for ACDC and 40\% for MMWHS), the respective semi-supervised setups asymptotically approach the fully supervised setup (differs by 0.3\% in ACDC and 1.1\% in MMWHS), thus depicting the efficacy of our framework under limited annotations.



\subsubsection{Comparison to state-of-the-art}

\begin{figure}
    \centering
    \includegraphics[scale = 0.5]{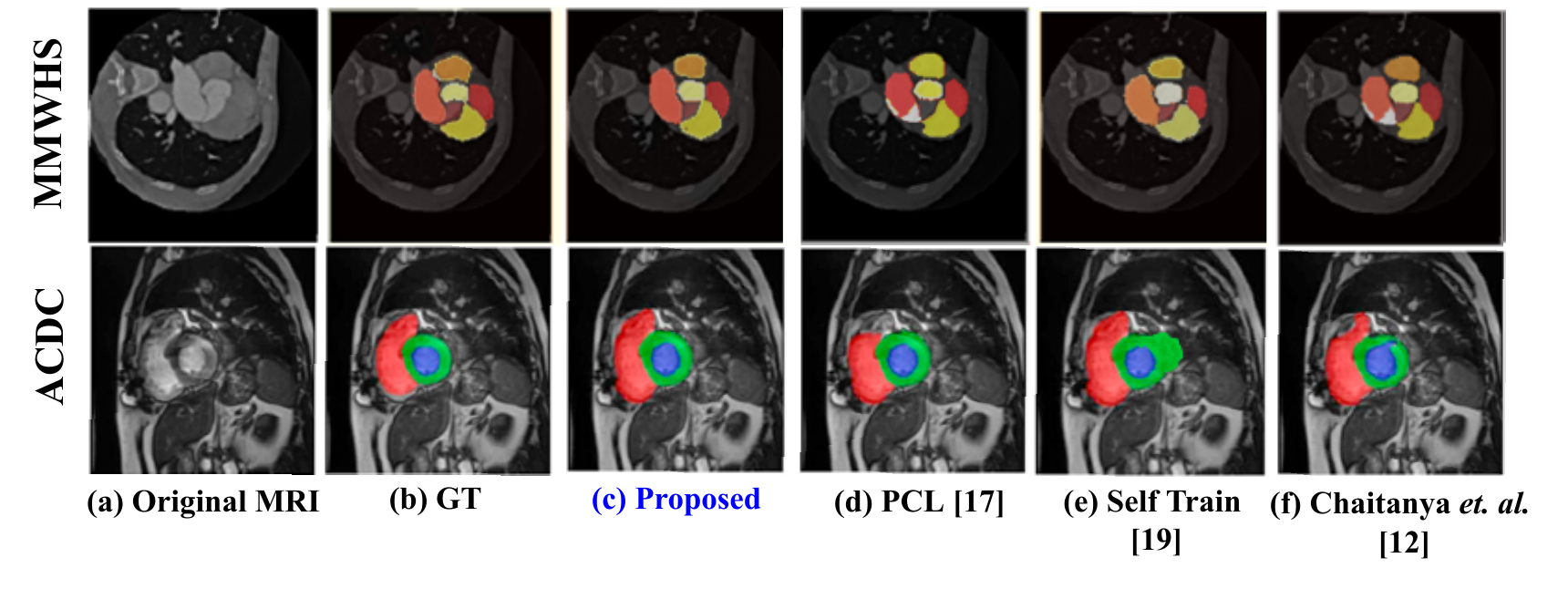}
    \caption{Visual comparison of our results with SoTA methods and ground truth, thus qualitatively validating the superiority of IDEAL in terms of segmentation performance.}
    \label{fig:segoutputs}
\end{figure}

We have also compared our proposed framework with several existing state-of-the-art semi-supervised cardiac MRI segmentation methods in literature \cite{chaitanya2020contrastive, chen2020simple, zeng2021positional, chen2019self, hu2021semi, chaitanya2019semi, bai2017semi, wu2022mutual}, the results of which are tabulated in Table \ref{tab:sotas}. Additionally, we also provide a visual comparison of our segmentation performances with the SoTA methods in Fig. \ref{fig:segoutputs}.

\begin{table}[t]
    \centering
    \caption{Performance Comparison (DSC scores) of the proposed IDEAL framework with SoTA methods in the literature on the ACDC and MMWHS datasets.} 
    \label{tab:sotas}
    \resizebox{\columnwidth}{!}{
    \begin{tabular}{l ccc ccc}

    \toprule
    \textbf{Method} & \multicolumn{3}{c}{\textbf{Average DSC (ACDC)}} & \multicolumn{3}{c}{\textbf{Average DSC (MMWHS)}}\\[0.5ex]
    \cmidrule(lr){2-4}
    \cmidrule(lr){5-7}
    & \textbf{L=1.25\%} & \textbf{L=2.5\%} & \textbf{L=10\%} & \textbf{L=10\%} & \textbf{L=20\%} & \textbf{L=40\%}\\[0.5ex]
    \midrule
    
    Chaitanya \etal \cite{chaitanya2020contrastive} & 0.725 & 0.789 & 0.872 & 0.569 & 0.694 & 0.794 \\[.5ex]

    Global CL \cite{chen2020simple} & 0.729 & \textbf{--} & 0.847 & 0.500 & 0.659 & 0.785 \\[.5ex]

    PCL \cite{zeng2021positional} & 0.671 & \textbf{0.850} & \textbf{0.885} & \textbf{--} & \textbf{--} & \textbf{--} \\[.5ex]

    Context Restoration \cite{chen2019self} & 0.625 & 0.714 & 0.851 & 0.482 & 0.654 & 0.783 \\[.5ex]
    
    MC-Net \cite{wu2022mutual} & 0.677 & 0.724 & 0.855 & 0.551 & 0.654  & 0.798 \\[.5ex]
    
    Label Efficient \cite{hu2021semi} & \textbf{--} & \textbf{--} & \textbf{--} & 0.382 & 0.553 & 0.764 \\[.5ex]

    Data Augmentation \cite{chaitanya2019semi} & 0.731 & 0.786 & 0.865 & 0.529 & 0.661 & 0.785 \\[.5ex]

    Self Train \cite{bai2017semi} & 0.690 & 0.749 & 0.860 & 0.563 & 0.691 & 0.801 \\[.5ex]

    \rowcolor{LightCyan}
    Ours & \textbf{0.738} & 0.846 & 0.879 & \textbf{0.626} & \textbf{0.791} & \textbf{0.815} \\[.5ex]
    
    \bottomrule
    
    \end{tabular} %
    }
    
\end{table}

\keypoint{ACDC Dataset} From Table \ref{tab:sotas}, it can be observed that our IDEAL model performs the best in $ L=1.25\%$ setting, while being highly competitive to  PCL \cite{zeng2021positional} in $L=2.5\% (\sim0.004)$ and $L=10\% (\sim0.006)$. This shows that even in severely constrained conditions, IDEAL can perform optimally.

\keypoint{MMWHS Dataset} Compared to existing state-of-the-art methods, our IDEAL model outperforms all previous frameworks by large margins, as inferred from Table \ref{tab:sotas}. Chaitanya \etal \cite{chaitanya2020contrastive} achieved the next best performance with DSC values of 56.9\% ($L=10\%$), 69.4\% ($\sim$10\% less than ours) with 20\% labeled data and 79.4\% with 40\% labeled data.

%
\section{Conclusion}
In this paper, we propose a semi-supervised approach for cardiac MRI segmentation to tackle the labeling bottleneck in medical imaging. Our framework leverages self-supervised contrastive learning with a novel projection head to capture dense local feature representation during pre-training, followed by employing a unique cross-consistency regularization scheme during model fine-tuning for the downstream segmentation task. Results show that our IDEAL framework outperforms several state-of-the-art methods on two widely used cardiac MRI datasets. We believe that such a paradigm can also be employed in cross-modal and cross-domain scenarios, as well as extended to natural images, something we intend to explore in near future.

\section{Acknowledgement}
This work was supported by the Basic Science Research Program through the National Research Foundation of Korea (NRF) funded by the Ministry of Education (2021R1I1A3049810).

\bibliographystyle{IEEEbib}
\bibliography{refs}

\end{document}